\documentclass{article}
\usepackage{ijcai26}

\usepackage{times}
\usepackage{soul}
\usepackage{url}
\usepackage[hidelinks]{hyperref}
\usepackage[utf8]{inputenc}
\usepackage[small]{caption}
\usepackage{graphicx}
\usepackage{amsmath}
\usepackage{amsthm, bm}
\usepackage{amssymb}
\usepackage{booktabs}
\usepackage{algorithm}
\usepackage{algorithmic}
\usepackage[switch]{lineno}
\usepackage{amssymb}
\usepackage{tabularx} 
\usepackage{multirow} 
\usepackage{subcaption}
\usepackage{enumitem}

\title{Explainable Knowledge Tracing \\via Probabilistic Embeddings and Pattern-based Reasoning}

\author{
Siyu Wu $^{1,2}$
\and
Cong Xu $^2$\and
Wei Zhang $^2*$\\
\affiliations
$^1$Shanghai Institute of AI Education, East China Normal University, Shanghai, 200241, China \\
$^2$Shool of Computer Sicence and Technology, East China Normal University, Shanghai, 200241, China\\
\emails
sywu04@gmail.com,
congxueric@gmail.com,
zhangwei.thu2011@gmail.com
}

\begin{document}

\maketitle

\begin{abstract}
Knowledge Tracing (KT) models students’ knowledge states based on learning interactions to predict performance.
While deep learning-based KT models have boosted predictive accuracy, most models rely on deterministic vector embeddings and opaque latent state transitions, limiting interpretability regarding how specific past behaviors influence predictions.
To address this limitation, we propose Probabilistic Logical Knowledge Tracing (PLKT), an interpretable KT framework that formulates prediction as a goal-conditioned evidence reasoning process over historical learning behaviors.
Instead of representing knowledge states as deterministic vector embeddings, PLKT employs robust Beta-distributed probabilistic embeddings to represent student knowledge states.
This probabilistic foundation allows us to model the uncertainty of historical behaviors and perform explicit logical operations (e.g., conjunction), constructing transparent reasoning paths that reveal how specific past interactions contribute to the prediction.
Extensive experiments show that PLKT outperforms state-of-the-art KT methods while achieving superior interpretability.
Our code is available at https://anonymous.4open.science/r/PLKT-D3CE/.
\end{abstract}

\section{Introduction}

Accurately modeling students’ evolving knowledge states is a fundamental requirement for personalized learning in intelligent education systems~\cite{wu2024comprehensive}.
Knowledge tracing (KT) predicts future performance from historical interactions and underpins applications such as exercise recommendation~\cite{wu2020exercise,wu2023contrastive}, learning path planning~\cite{chen2023knowledge}, and instructional intervention~\cite{yu2025exploring}.
Recent deep learning–based KT (DLKT) models have achieved substantial performance gains~\cite{piech2015deep,zhang2017dynamic,ghosh2020context,choi2020towards,xia2025flatformer}.
However, these improvements often come at the cost of interpretability: predictions are produced through high-dimensional latent representations, making it difficult to explain \emph{which} historical behaviors contribute to a prediction and \emph{why} a particular outcome is inferred.
Such opacity limits their application in educational settings, where teachers and students need interpretable feedback. 
Moreover, most methods adopt deterministic vector embeddings, ignoring uncertainty in students’ mastery and potentially yielding overconfident or unstable predictions~\cite{cheng2025uncertainty}.

\begin{figure}
    \centering
    \includegraphics[width=\linewidth]{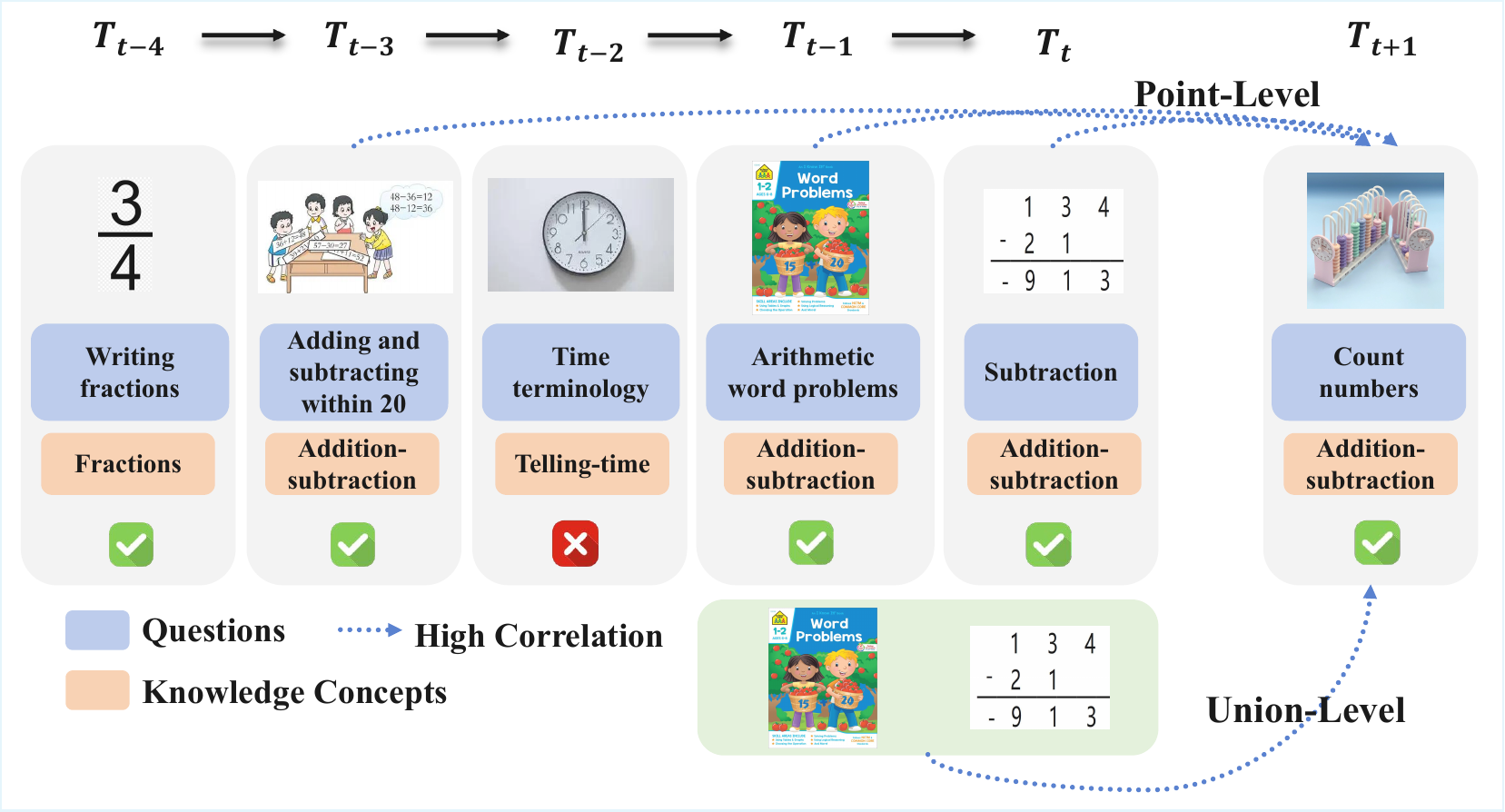}
    \vspace{-1em}\caption{Illustration of interpretable knowledge tracing via point-wise and union-wise patterns.}
    \label{fig:intro_case}
\end{figure}

To improve interpretability in knowledge tracing, prior work has explored diverse directions, including incorporating explicit knowledge structures \cite{shen2025enhancing}, modeling fine-grained concept relations \cite{duan2024towards}, introducing causal reasoning \cite{minn2022interpretable}, leveraging large language models \cite{li2025cikt}, and embedding psychometric principles such as Item Response Theory \cite{yeung2019deep,sun2024interpretable}.
To address uncertainty, Cheng et al. proposed the UKT model \cite{cheng2025uncertainty}, which characterizes the uncertainty in students' knowledge states through stochastic probabilistic embeddings.
Despite these advances, these approaches still suffer from three key limitations:
(1) First, although many KT models can aggregate information from multiple historical interactions, such aggregation is typically performed implicitly in latent spaces (e.g., via recurrent neural networks or attention). 
Their explanations still rely on isolated interactions, without explicitly representing joint learning evidence formed through continuous practice.
As shown in Figure \ref{fig:intro_case}, when predicting a student’s response to a target question, the combined pattern of ‘Arithmetic word problems’ and ‘Subtraction’ more robustly represents students' sustained consolidation and stable mastery of the target concept than point-level conceptual association questions.
While existing models may implicitly exploit such information, they do not expose these joint behavioral segments as interpretable reasoning units, thereby limiting transparency in explanations.
(2) Second, most interpretability mechanisms rely on latent attention weights or embedding similarities, lacking explicit semantic or structural grounding.
(3) Third, although probabilistic representations have been introduced to model uncertainty, they are seldom coupled with structured reasoning, limiting their ability to support transparent and coherent decision processes.

To address these challenges, we propose Probabilistic Logical Knowledge Tracing (PLKT), which reframes KT from sequential state prediction to pattern-based evidence reasoning.
PLKT represents knowledge states using Beta distribution embeddings, replacing point vectors with probabilistic representations that explicitly capture uncertainty and naturally support probabilistic operations.
It further extracts multi-level behavioral patterns, which serve as interpretable evidence units.
For different patterns, logical conjunctions are employed to aggregate evidence and determine pattern contributions, making the prediction process transparent and auditable.
This design yields transparent, goal-conditioned predictions grounded in explicit learning patterns.

The main contributions of this paper are summarized as:
\begin{itemize}
    \item We propose PLKT, the first knowledge tracing model that unifies probabilistic embeddings, multi-level behavioral patterns, and pattern-based reasoning within a single interpretable architecture.
    \item We introduce Beta-distributed probabilistic embeddings to model students’ knowledge states, and perform explicit logical operations for aggregation, enabling structured and interpretable pattern-based reasoning.
    \item We design explicit pattern contribution mechanisms that quantify how specific historical learning behaviors support each prediction, yielding transparent and cognitively grounded reasoning pathways.
    \item Extensive experiments across five publicly available educational datasets demonstrate that PLKT achieves state‑of‑the‑art or highly competitive prediction performance while offering superior interpretability.
\end{itemize}









\section{Related Works}
\label{sec:related_work}
\subsection{Explainable Knowledge Tracing}
Knowledge tracking (KT) is a fundamental task in personalized learning \cite{shen2024survey}, where interpretability holds equal importance to predictive accuracy, thereby driving extensive research into interpretable KT.
Existing interpretable KT methods can be broadly categorized into ante-hoc and post-hoc approaches.
Ante-hoc methods embed interpretability into model design, including classical probabilistic models such as BKT and its extensions \cite{nedungadi2014predicting}, as well as neural KT models with interpretable components.
Representative examples include attention-based methods (e.g., SAKT \cite{pandey2019self}, SAINT \cite{choi2020towards}, AKT \cite{ghosh2020context}), which highlight influential historical interactions via attention weights \cite{liu2023simplekt,pandey2020rkt}, and theory-driven models \cite{su2021time,zhang2017dynamic,gan2020knowledge,fosnot2013constructivism}  grounded in educational or psychometric principles, such as Deep-IRT \cite{yeung2019deep} and PKT \cite{sun2024progressive}.
Post-hoc approaches interpret trained models through latent attribution or visualization without modifying model structure, with recent work incorporating causal inference \cite{huang2024learning,zhou2025disentangled} or ensemble frameworks \cite{sun2022ensemble}.
Furthermore,  large language models (LLMs) have been explored to generate interpretable textual representations for KT via semantic profiling or concept extraction \cite{duan2025automated,li2025cikt}.

Despite these advances, most KT models provide interpretability mainly through latent weights or local attributions, which lack stable semantic grounding.
Reasoning generated by LLMs may fail to authentically reflect decision-making processes, leading to inconsistencies. 
Furthermore, while sequential models can implicitly aggregate historical information, their reasoning remains opaque, hindering traceable and cognitively meaningful explanations.
In contrast, PLKT employs probabilistic embeddings and explicit logical evidence aggregation, achieving both strong predictive performance and structured interpretability.

\subsection{Embedding Methods for Education}
Embedding techniques are widely used in educational data mining, with most KT models relying on one-hot or ID embeddings to represent students, questions, and knowledge concepts \cite{piech2015deep,zhang2017dynamic}.
However, such vector embeddings inadequately capture the uncertainty and structural complexity of students’ knowledge states, motivating the exploration of more expressive representations.
In general online learning scenarios, PHE \cite{li2025probabilistic} models categorical embeddings as probability distributions and updates them via Bayesian online learning, demonstrating robustness to emerging categories.
Within the knowledge tracing domain, UKT \cite{cheng2025uncertainty} is among the few works that explicitly adopt probabilistic embeddings.
UKT represents knowledge states as distributions and employs Wasserstein-distance-based self-attention to model the evolution of student learning uncertainty.
These studies demonstrate the potential of probabilistic representations for capturing uncertainty in KT and suggest a promising direction for more interpretable knowledge state modeling.
Building on these insights, we employ beta distribution embeddings to model student knowledge states, enabling pattern-based reasoning and pattern-level explanations.

\begin{figure*}
    \centering
     \includegraphics[width=16.8cm]{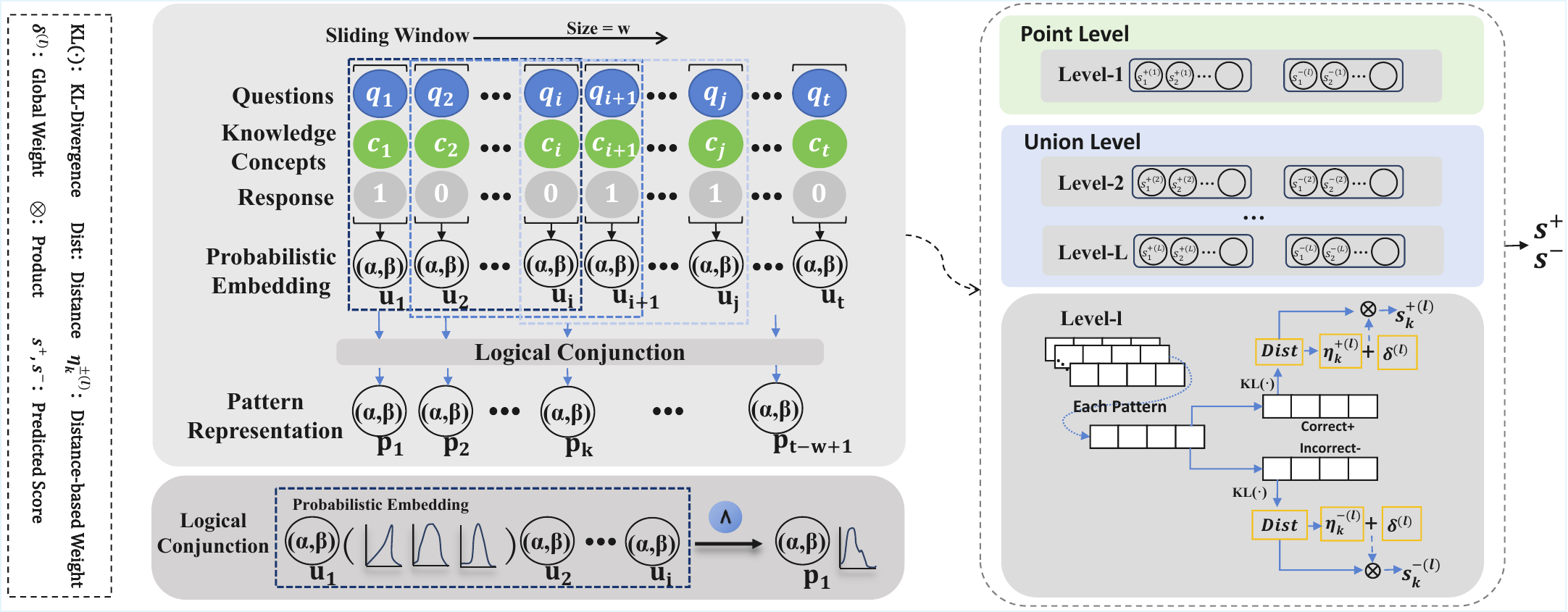}
    \vspace{-1em}\caption{The overview of the proposed PLKT framework. The right side of the figure shows the level-wise prediction scores ($s_k^{+(l)},s_k^{-(l)}$). The detailed computation is illustrated only for level $l$, while all other levels share the same scoring mechanism.}
    \label{PLKT_model}
\end{figure*}

\section{Preliminary}
\subsection{Problem Formulation}
\label{sec:prob_formu}
Knowledge Tracing (KT) models students’ evolving knowledge states from their learning history to predict future performance.
Formally, let $\mathcal{U}$, $\mathcal{Q}$, and $\mathcal{C}$ denote the sets of students, questions, and knowledge concepts.
For a student $u \in \mathcal{U}$, the learning history is represented as a time-ordered interaction sequence $X^{u}_{1:t}=\{(q^{u}_1,c^{u}_1,r^{u}_1),\ldots,(q^{u}_t,c^{u}_t,r^{u}_t)\}$, where $q^u_i \in \mathcal{Q}$ denotes the question attempted at time $i$, $c^u_i \in \mathcal{C}$ is the associated knowledge concept, and $r^u_i \in \{0,1\}$ indicates the response outcome, with $r_i^u=1$ denoting a correct answer.
For clarity, this paper assumes each question relates to a single knowledge concept, but this setting can be directly extended to scenarios involving multiple concepts.
When the context is clear, the student identifier $u$ is omitted.
At time step $t$, given the historical interaction sequence $X_{1:t}$, the student encounters a target question $q_{t+1}$ associated with the concept $c_{t+1}$.
The objective of KT is to estimate the conditional probability $P(r_{t+1}=1 \mid X_{1:t}, q_{t+1}, c_{t+1})$ that a student will correctly answer question $q_{t+1}$.
Other notations are summarized in Supplementary Material A.
\subsection{Logical Conjunction}
\label{sec:beta_emb}

In this part, we will introduce the Beta distribution and the logical conjunction operation in probability theory, which will be applied in subsequent questions and the representation of knowledge concepts.
Recall that a Beta distribution $\text{Beta}(\alpha, \beta)$ is parameterized by two positive shape parameters $\alpha, \beta \in \mathbb{R}_+$. 
A $d$-dimensional Beta embedding can thus be viewed as a collection of $d$ independent Beta distributions~\cite{ren2020beta}, 
denoted by $\text{Beta}(\bm{\alpha}, \bm{\beta})$, where $\bm{\alpha}, \bm{\beta} \in \mathbb{R}_+^d$.
Compared with conventional vector embeddings, Beta embeddings provide principled probabilistic semantics and support well-defined distributional operations, making them particularly suitable for interpretable reasoning in knowledge tracing. The probabilistic conjunction operator `$\bigwedge$' over $n$ Beta embeddings $\{\text{Beta}(\bm{\alpha}_i, \bm{\beta}_i)\}_{i=1}^n$ produces a fused embedding defined as
\begin{align}
\label{eq:log_conju}
\bigwedge_{i=1}^n \text{Beta}(\bm{\alpha}_i, \bm{\beta}_i)
= \text{Beta}\left(
\sum_{i=1}^n \bm{w}_i \odot \bm{\alpha}_i,
\sum_{i=1}^n \bm{w}_i \odot \bm{\beta}_i
\right),
\end{align}
where $\odot$ denotes the Hadamard product. The importance weights $\bm{w}_i$ are computed via a self-attention mechanism:
\begin{align*}
\bm{w}_i
= \frac{
\exp\left(\text{MLP}([\bm{\alpha}_i; \bm{\beta}_i])\right)
}{
\sum_{j=1}^n \exp\left(\text{MLP}([\bm{\alpha}_j; \bm{\beta}_j])\right)
}, \quad i = 1, 2, \ldots, n.
\end{align*}
Here, $\text{MLP}(\cdot)$ is a learnable function introduced to enhance the expressive capacity of the model. Unless otherwise specified, the MLPs used in different components are independently parameterized.

\section{The Architecture of PLKT}
\label{sec:model}

\subsection{Feature Modeling Module}
\label{sec_mode_inter}
Given a student’s interaction sequence $X_{1:t}$, we explicitly differentiate learning signals from correct and incorrect responses.
Therefore, the original question and knowledge concept indices are offset based on the response $r_i$.
Subsequently, both the question and knowledge concept are embedded into Beta distributions via a learnable embedding function:
\begin{equation}
(\bm{\alpha}_i^q,\bm{\beta}_i^q)=h_{\mathrm{emb}}^q(q_i,r_i), \quad
(\bm{\alpha}_i^c,\bm{\beta}_i^c)=h_{\mathrm{emb}}^c(c_i,r_i),
\label{eq:base_beta}
\end{equation}
where $h_{\text{emb}}^q, h_{\text{emb}}^c$: $\mathbb{N} \to \mathbb{R}^{2d}$, and each output is split into $\bm{\alpha},\bm{\beta} \in \mathbb{R}^d$.
Compared to vector embeddings, Beta distributions explicitly model mastery and uncertainty, enhancing the model's expressive power~\cite{xiong2023geometric}.

Beyond the question of identity and knowledge concept mastery, question difficulty offers an informative prior for modeling student performance.
Following classical educational measurement theory and prior work~\cite{huang2017question,zhang2021multi}, we estimate question difficulty using empirical accuracy statistics.
For a question $q$, its accuracy rate is $\hat{a}_q = \frac{M_q}{N_q}$, where $N_q$ and $M_q$ denote the number of attempts and correct responses, respectively, and the difficulty is defined as $d_q = 1 - \hat{a}_q \in [0,1]$.
Unobserved questions are assigned a neutral prior $d_q=0.5$.
Given the empirical difficulty score $d_{q_i}$ , difficulty-aware modulation is applied to the question-level distribution:
\begin{equation}
\label{eq:difficulty_modulation}
\begin{aligned}
\bm{\alpha}_i^q & \leftarrow \mathrm{Softplus}(\bm{\alpha}_i^q + \mathbf W_\alpha d_{q_i}) + \epsilon, \\
\bm{\beta}_i^q & \leftarrow \mathrm{Softplus}(\bm{\beta}_i^q + \mathbf W_\beta d_{q_i}) + \epsilon,
\end{aligned}
\end{equation}
where $\mathbf{W}_{\alpha},\mathbf{W}_{\beta} \in \mathbb{R}^{d}$ are learnable parameters.
The Softplus activation ensures that the shape parameters of the Beta distribution remain strictly positive, while $\epsilon$ is a small constant added for numerical stability.

Since a student's response is jointly determined by knowledge concept mastery and question characteristics, we unify the concept‑level and question‑level Beta embeddings via the probabilistic conjunction operator (Eq.~\ref{eq:log_conju}):
$(\bm{\alpha}_i,\bm{\beta}_i)
= (\bm{\alpha}_i^c,\bm{\beta}_i^c)\ \wedge\ (\bm{\alpha}_i^q,\bm{\beta}_i^q)$,
yielding a unified probabilistic representation for interaction $i$.
For the target question $q_{t+1}$ and its associated concept $c_{t+1}$, we apply the same response-aware embedding, difficulty modulation, and conjunction mechanism to construct two target distributions corresponding to correct and incorrect response states, denoted as $(\bm{\alpha}^+_{t+1},\bm{\beta}^+_{t+1})$ and $(\bm{\alpha}^-_{t+1},\bm{\beta}^-_{t+1})$.
These target distributions are semantically aligned with historical interaction representations and serve as reference states for pattern–target matching.



\subsection{Multi-level Pattern Extraction Module}
\label{sec:multi_level_pattern}
Most existing DLKT models aggregate historical interactions via recurrent neural networks or attention-based mechanisms \cite{piech2015deep,ghosh2020context}, but their reasoning processes remain embedded in latent states or attention weights, limiting interpretability.
From an explanatory perspective, learning behaviors often form semantically coherent segments across temporal scales, whose combined outcomes provide more cognitively meaningful evidence than isolated interactions.
To this end, inspired by research in the recommended field  \cite{ma2025pattern}, we propose a multi-level pattern extraction mechanism to explicitly capture learning behaviors at different temporal granularities.

Multi-level patterns are extracted from ${(\bm{\alpha}_i,\bm{\beta}_i)}_{i=1}^{t}$ using sliding windows of lengths $l\in\{1,\dots,L\}$ with stride 1, where the $k$-th pattern at level $l$ is defined as
$\mathcal{P}_k^{(l)}=\left[(\bm{\alpha}_k,\bm{\beta}_k),(\bm{\alpha}_{k+1},\bm{\beta}_{k+1}),\dots,(\bm{\alpha}_{k+l-1},\bm{\beta}_{k+l-1})\right]$,
$k\in\{1,\dots,t-l+1\}$.
Patterns at different levels convey distinct semantics.
When $l=1$, $p^{(1)}$ reduces to a point-level interaction, providing direct and intuitive evidence of individual behavioral influence.
For $l>1$, patterns represent union-level structures formed by consecutive interactions that reflect a more coherent learning context than isolated responses, yielding more interpretable explanatory units.
Accordingly, patterns from all levels of $1$ to $L$ are retained, and the remaining challenge lies in aggregating these patterns into unified representations for transparent decision-making.

To obtain a unified probabilistic representation for each pattern, we aggregate the constituent interaction distributions using the probabilistic conjunction operator defined in Eq.~\ref{eq:log_conju} along the temporal dimension:
\begin{equation}
\label{eq:pattern_conjunction}
\mathbf{p}_k^{(l)} :=(\bm{\alpha}_k^{(l)}, \bm{\beta}_k^{(l)}) = \bigwedge_{i=1}^{l} \text{Beta} (\bm{\alpha}_{k+i-1}, \bm{\beta}_{k+i-1}),
\end{equation}
where $l$ denotes the pattern length.

Unlike an interaction-level conjunction that combines multiple attributes within a single event, this operator aggregates temporally distributed learning evidence, yielding a coherent pattern-level representation.
By modeling both point-level and union-level patterns, this multi-level representation captures short-term fluctuations and longer-term learning trends, providing stable and interpretable behavioral pattern representations for subsequent pattern–target matching.

\subsection{Pattern Contribution Aggregation Module}
\label{sec:pattern_agg}
This module estimates the contribution of each historical learning pattern to the prediction by jointly considering its local alignment with the target question and its global importance within the learning trajectory.

\subsubsection{Local Pattern-target Matching}
Since both historical patterns and target response states are modeled as Beta distributions, we quantify their alignment using Kullback–Leibler (KL) divergence, which captures discrepancies in both expected mastery and uncertainty.
Compared to Euclidean or cosine distance, KL divergence provides a more expressive similarity measure for probabilistic knowledge representations.

Given a pattern $\mathbf{p}_k^{(l)}$ at level $l$, we compute its distance to the correct and incorrect target representations:
\begin{equation*}
\label{eq:pattern_target_distance}
\begin{aligned}
D_k^{\pm(l)} &= \sum_{j=1}^{d} \text{KL}(\text{Beta}(\bm{\alpha}_{t+1,j}^\pm, \bm{\beta}_{t+1,j}^\pm) \parallel \text{Beta}(\bm{\alpha}_{k,j}^{(l)}, \bm{\beta}_{k,j}^{(l)})). \\
\end{aligned}
\end{equation*}
Smaller divergence values indicate stronger alignment between the pattern and the corresponding target state.
However, directly using KL divergence as weights is suboptimal, as absolute distances tend to be large during early training and gradually shrink as representations improve, making it difficult to highlight relatively important patterns.
We therefore transform divergences into similarity scores and normalize them into attention weights.
Specifically, similarity scores are defined as $s_k^{\pm(l)} = \gamma - D_k^{\pm(l)}$,where $\gamma>0$ is a learnable temperature parameter.
The final distance-based attention weights are obtained via Softmax normalization within each pattern level:
\begin{equation*}
\label{eq:distance_based_weights}
\eta_k^{+(l)} = \frac{\exp(s_k^{+(l)})} {\sum_{j=1}^{t-l+1} \exp(s_j^{+(l)})},\quad
\eta_k^{-(l)} =\frac{\exp(s_k^{-(l)})}{\sum_{j=1}^{t-l+1} \exp(s_j^{-(l)})}\,.
\end{equation*}

This matching mechanism assigns larger weights to patterns that are more consistent with the corresponding target response state.
Normalization is performed independently at each pattern level, allowing patterns at the same temporal scale to compete, while cross-level interactions are handled in the subsequent aggregation module.

\subsubsection{Global Pattern Importance Modeling}
\label{sec:pattern_agg}
Local weights are computed independently for each pattern and do not explicitly capture the global learning trajectory reflected in the entire interaction sequence.
Therefore, we introduce a learnable global importance modeling component to complement local matching.


For each interaction $i$, the Beta-distributed knowledge state $(\bm{\alpha}_i,\bm{\beta}_i)$ is summarized by its expectation
$\mathbf{e}_i=\frac{\bm{\alpha}_i}{\bm{\alpha}_i+\bm{\beta}_i}\in\mathbb{R}^d$.
Stacking all interactions yields a sequence-level representation
$\mathbf{E}=[\mathbf{e}_1;\dots;\mathbf{e}_t]\in\mathbb{R}^{t\times d}$.
Flattened sequence representations are fed into a multilayer perceptron to produce level-specific importance scores:
\begin{equation}
\label{eq:pattern_weight_mlp}
\boldsymbol{\delta}^{(l)}
= \mathrm{Softmax}\left(
\mathrm{MLP}\left(\mathrm{Flatten}(\mathbf{E})\right)
\right)
\in \mathbb{R}^{t-l+1},
\end{equation}
where each $\delta_k^{(l)}$ reflects the contribution of pattern $p_k^{(l)}$ from a global, sequence-aware perspective.
Independent weighting networks are employed for different levels due to varying pattern counts.

Finally, the distance-based and the learnable weights are linearly fused to obtain the final pattern contributions:
\begin{equation}
\label{eq:weight_fusion}
w_k^{+(l)} = \eta_k^{+(l)} + \lambda \cdot \delta_k^{(l)}, \quad
w_k^{-(l)} = \eta_k^{-(l)} + \lambda \cdot \delta_k^{(l)},
\end{equation}
where $\lambda\in[0,1]$ is used to adjust the relative influence of the global learnable weights.

This aggregation strategy yields interpretable pattern contributions by jointly considering
\emph{how well} each pattern matches the target state and
\emph{how important} it is within the overall learning trajectory,
providing a principled and transparent foundation for subsequent prediction.

\subsection{Prediction and Training}
KT aims to predict whether a learner will answer the target question correctly, rather than only modeling the question itself.
Accordingly, PLKT adopts an outcome-conditioned evidence aggregation scheme, where historical patterns contribute separately to the \emph{correct} and \emph{incorrect} target states based on their estimated contributions.
Specifically, for each outcome, PLKT first aggregates matching scores within each pattern length $l$, followed by evidence integration across multiple pattern levels:
\begin{equation}
\label{eq:cross_level_aggregation}
\begin{aligned}
s^\pm  &= \sum_{l=1}^{L} s^{\pm(l)}, \quad
s^{\pm(l)} = \sum_{k=1}^{t-l+1} w_k^{\pm(l)} \cdot s_k^{\pm(l)}.
\end{aligned}
\end{equation}
The probability of correctly answering $q_{t+1}$ is predicted as
\begin{equation}
\label{eq:probability_prediction}
\hat{r}_{t+1}
= P(r_{t+1}=1 \mid X_{1:t}, q_{t+1}, c_{t+1})
= \sigma(s^+ - s^-),
\end{equation}
where $\sigma(\cdot)$ is the sigmoid function.

PLKT is trained using the binary cross-entropy loss:
\begin{equation}
\label{eq:training_objective}
\mathcal{L}
= -\frac{1}{N} \sum_{i=1}^{N}
\left[
r_i \log \hat{r}_i + (1-r_i)\log(1-\hat{r}_i)
\right],
\end{equation}
where $r_i \in \{0,1\}$ is the ground-truth response.
By jointly optimizing all parameters, PLKT integrates multi-level pattern evidence with probabilistic reasoning, achieving accurate and interpretable knowledge state prediction.

\section{Experiments}
This section presents a comprehensive evaluation of PLKT through extensive experiments, including baseline comparisons, analysis of key component contributions, and multi-level pattern analysis.
Its interpretability is further examined with relevant case studies. Additionally, hyperparameter analysis is detailed in Supplementary Material B.5 due to space constraints.

\begin{table}[!t]
  \centering
  \footnotesize
  \setlength{\tabcolsep}{4pt} 
  \resizebox{0.48\textwidth}{!}{
  \begin{tabular}{lccccc}
    \toprule	
    Dataset  & ASSIST09 & ASSIST12 & Junyi & Algebra05 & Bridge06 \\
    \midrule 
    \#Stu      & 4.2K   & 27.1K  & 87.5K  & 0.6K    & 1.1K   \\
    \#Q     & 17.7K  & 51.0K  & 705    & 173.1K  & 129.3K \\
    \#KC      & 123    & 245    & 39     & 112     & 493    \\
    \#Int  & 458.8K & 2.63M  & 500K   & 607.0K  & 1.82M  \\     
    \bottomrule
  \end{tabular}}
  \vspace{-1em}\caption{Dataset Information.}
   \label{tab:dataset}
\end{table}

\subsection{Experimental Setup}
\noindent\textbf{Datasets.}
To comprehensively evaluate the effectiveness and generalization ability of PLKT, we conduct experiments on five widely used real-world knowledge tracing benchmark datasets: ASSIST09, ASSIST12, Junyi, Algebra2005, and Bridge2006.
The key statistics of the processed datasets are summarized in Table~\ref{tab:dataset}.
Detailed descriptions of each dataset are provided in Supplementary Material B.1.

\begin{table*}[htbp]
\centering
\small
\begin{tabularx}{0.95\textwidth}{@{}c *{10}{>{\centering\arraybackslash}X} @{}}
\toprule
\multicolumn{1}{c}{\multirow{2}{*}{Model}} & \multicolumn{2}{c}{ASSIST09}      & \multicolumn{2}{c}{ASSIST12}      & \multicolumn{2}{c}{Junyi}     & \multicolumn{2}{c}{Algebra05}   & \multicolumn{2}{c}{Bridge06} \\ \cmidrule(lr) {2-3}  \cmidrule(lr) {4-5} \cmidrule(lr) {6-7} \cmidrule(lr) {8-9} \cmidrule(lr) {10-11}
\multicolumn{1}{c}{}                       & ACC             & AUC             & ACC             & AUC             & ACC             & AUC             & ACC             & AUC             & ACC            & AUC           \\ \midrule
\multicolumn{1}{c|}{DKT}                   & 0.7346          & 0.7403          & 0.7267          & 0.7103          & 0.8297          & 0.6414          & 0.8617          & 0.9103          & 0.8485         & 0.7687        \\
\multicolumn{1}{c|}{DKT+}                  & 0.7349          & 0.7404          & 0.7269          & 0.7104          & 0.8299          & 0.6417          & 0.8614          & 0.9099          & 0.8489         & 0.7706        \\
\multicolumn{1}{c|}{DKVMN}                 & 0.7270          & 0.7278          & 0.7203          & 0.6930          & 0.8299          & 0.6417          & 0.8550          & 0.8994          & 0.8462         & 0.7555        \\
\multicolumn{1}{c|}{SAKT}                  & 0.7288          & 0.7309          & 0.7209          & 0.6915          & 0.8302          & 0.6435          & 0.8608          & 0.9084          & 0.8464         & 0.7572        \\
\multicolumn{1}{c|}{SAINT}                 & 0.7256          & 0.7141          & 0.7276          & 0.7045          & 0.8323          & 0.7067          & 0.8642          & 0.9141          & 0.8445         & 0.7609        \\
\multicolumn{1}{c|}{AKT}                   & {\underline {0.7455}}    & {\underline {0.7754}}    & 0.7526          & 0.7685          & 0.8319          & 0.7159          & 0.8699          & 0.9230          & {\underline {0.8560}}   & {\underline {0.8087}}  \\
\multicolumn{1}{c|}{SimpleKT}              & 0.7435          & 0.7702          & 0.7492          & 0.7601          & 0.8320          & 0.7170          & 0.8728          & 0.9262          & 0.8550         & 0.8076        \\
\multicolumn{1}{c|}{DeepIRT}               & 0.7275          & 0.7273          & 0.7213          & 0.6945          & 0.8299          & 0.6424          & 0.8552          & 0.8950          & 0.8448         & 0.7547        \\
\multicolumn{1}{c|}{CSKT}                  & 0.7429          & 0.7682          & 0.7462          & 0.7603          & 0.8307          & 0.7133          & 0.8702          & 0.9239          & 0.8546         & 0.8045        \\
\multicolumn{1}{c|}{LPKT}                  & 0.7415          & 0.7676          & {\underline {0.7528}}    & {\underline {0.7687}}    & {\underline {0.8328}}    & {\underline {0.7172}}    & 0.8109          & 0.8197          & 0.8530         & 0.7965        \\
\multicolumn{1}{c|}{SparseKT}              & 0.7388          & 0.7606          & 0.7459          & 0.7545          & 0.8324          & 0.7139          & 0.8686          & 0.9210          & 0.8510         & 0.7926        \\
\multicolumn{1}{c|}{UKT}                   & 0.7356          & 0.7418          & 0.7514          & {\underline {0.7687}}    & 0.8313          & 0.7157          & {\underline {0.8734}}    & {\underline {0.9281}}    & 0.8552         & 0.8085        \\ \midrule
\multicolumn{1}{c|}{PLKT}                 & \textbf{0.7657} & \textbf{0.8116} & \textbf{0.7613} & \textbf{0.7849} & \textbf{0.8333} & \textbf{0.7226} & \textbf{0.9072} & \textbf{0.9645} & \textbf{0.8653}               & \textbf{0.8534}              \\ \midrule
\multicolumn{1}{c|}{Improv.}               & 2.64\%           & 4.67\%           & 1.12\%           & 2.06\%           & 0.06\%           & 0.75\%           & 3.73\%           & 3.77\%           &  1.07\%              & 5.24\%           \\
\bottomrule
\end{tabularx}
\vspace{-1em}\caption{Performance comparison, where the best performance is highlighted in bold and the second-best is underlined.}
\label{tab:model_performance}
\end{table*}

\noindent\textbf{Baselines.}
To evaluate the effectiveness of the proposed PLKT model, we compare it with twelve representative KT baselines published between 2015 and 2025.
Classical approaches include DKT \cite{piech2015deep} and DKT+ \cite{yeung2018addressing}. 
DKVMN is adopted to represent memory-based methods \cite{zhang2017dynamic}. 
Attention-based and Transformer-style models include SAKT \cite{pandey2019self}, AKT \cite{ghosh2020context}, SAINT \cite{choi2020towards}, and SimpleKT \cite{liu2023simplekt}. 
Psychometric-inspired and interpretable models include DeepIRT \cite{yeung2019deep} and LPKT \cite{shen2021learning}. 
Recent models addressing data sparsity, cold-start, and uncertainty include SparseKT \cite{huang2023towards}, CSKT \cite{bai2025cskt}, and UKT \cite{cheng2025uncertainty}. 
Detailed descriptions of all baselines are provided in Supplementary Material B.2.

\noindent\textbf{Experimental Settings and Metrics.}
For fair comparison, all baselines are reproduced using the pyKT framework~\cite{liu2022pykt} with unified preprocessing and evaluation protocols, and their hyperparameters follow the original papers or pyKT defaults.
Models were trained using the Adam optimizer \cite{kinga2015method}.
The batch size was set to 256 for Junyi and 512 for the other datasets.
For PLKT, the learning rate was selected from $\{1e-4,2e-4,1e-3\}$ based on validation performance, with a fixed hidden dimension of 256 and dropout chosen from $\{0.1, 0.2, 0.5\}$.
The pattern layer number was set to 6 for Bridge2006, 3 for Junyi, and 5 for the remaining datasets.
Student interaction sequences were segmented into subsequences of length 80, with short or incomplete records removed and zero padding applied when necessary.
The data were split into training, validation, and test sets with a 7:1:2 ratio.
Since knowledge tracing is formulated as a binary classification task, model performance was evaluated using Area Under the ROC Curve (AUC) and Accuracy (ACC).
To ensure reliable results, all experiments were conducted with five-fold cross-validation.

\begin{table}[]
\centering
\resizebox{0.45\textwidth}{!}{
\begin{tabular}{lc|cccc}
\toprule
\multicolumn{2}{c|}{Data \& Metric}      & PLKT   & w/o Inter & w/o LW & w/o PE \\  \midrule
\multirow{2}{*}{ASSIST09}  & ACC & 0.7657 & 0.7576     & 0.7524            & 0.6531             \\
                           & AUC & 0.8116 & 0.7922     & 0.7949            & 0.7008             \\ 
\multirow{2}{*}{ASSIST12}  & ACC & 0.7613 & 0.7566     & 0.7512            & 0.6907             \\
                           & AUC & 0.7849 & 0.7747     & 0.7658            & 0.6113             \\ 
\multirow{2}{*}{Junyi}     & ACC & 0.8333 & 0.8325     & 0.8311            & 0.8228             \\
                           & AUC & 0.7226 & 0.7181     & 0.7031            & 0.6907             \\ 
\multirow{2}{*}{Algebra05} & ACC & 0.9072 & 0.8948     & 0.8611            & 0.8312             \\
                           & AUC & 0.9645 & 0.9492     & 0.9233            & 0.8701             \\ 
\multirow{2}{*}{Bridge06}  & ACC & 0.8653 & 0.8608     & 0.8615            & 0.8236             \\
                           & AUC & 0.8534 & 0.8311     & 0.8369            & 0.8141             \\ \bottomrule
\end{tabular}}
\vspace{-1em}\caption{Results of the ablation experiments.}
\label{tab:ablation_study}
\end{table}

\subsection{Results}
\subsubsection{Overall Performance}
Table~\ref{tab:model_performance} reports the performance comparison between PLKT and representative KT baselines on five public datasets.
Overall, PLKT consistently achieves the best results across all datasets.
Compared with the strongest baseline on each dataset, PLKT attains relative improvements of 0.06\%–3.73\% in ACC and 0.75\%–5.24\% in AUC, demonstrating clear and robust superiority.
PLKT exhibits particularly pronounced gains on complex datasets such as Bridge2006 and Algebra2005, which involve large concept spaces, intricate interactive relationships, and long learning sequences.
In such settings, point-wise or shallow sequence models struggle to capture higher-order behavioral structures.
By explicitly constructing joint learning patterns and performing pattern-based reasoning, PLKT achieves sustained performance gains through the high-level integration of historical evidence.

On ASSIST09, PLKT also achieves stable improvements, indicating that multi-level pattern modeling remains beneficial even at moderate data scales.
On Junyi, which exhibits structural simplicity (single-concept questions and short sequences), the gains are smaller (+0.06\% ACC, +0.75\% AUC), yet PLKT still consistently outperforms all baselines, underscoring its robustness across datasets of varying complexity.
Among baselines, attention-based models consistently outperform earlier deep learning methods, highlighting the importance of modeling interaction dependencies.
Interpretable models such as DeepIRT tend to trade predictive accuracy for explainability.
In contrast, PLKT alleviates this trade-off by leveraging probabilistic pattern representations, achieving superior performance while maintaining interpretability.
Overall, these results confirm the effectiveness of PLKT in capturing complex learning behaviors through principled pattern matching and aggregation.

\subsubsection{Ablation Study for Different Components}
\label{sec:ablation_study}

\begin{figure}
    \centering
    \includegraphics[width=\linewidth]{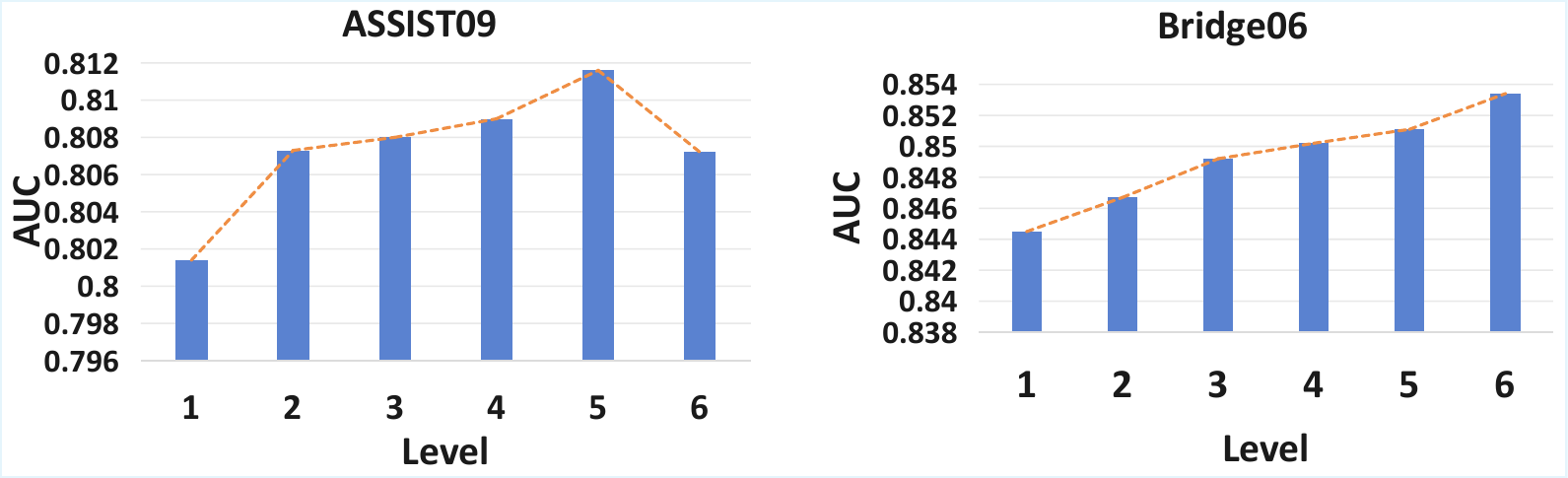}
    \vspace{-1em}\caption{AUC performance of PLKT with varying pattern-levels on ASSIST09 and Bridge06. $level=1$ denotes point level, while $level>1$ indicates simultaneous use of multiple levels.}
    \label{fig:pattern_level_09_12}
\end{figure}

Ablation studies of each core component in PLKT are exhibited in Table~\ref{tab:ablation_study}.
Specifically, we evaluate the following variants:
    (1) w/o Inter: Replaces the probabilistic logical conjunction between questions and knowledge concepts with simple vector concatenation.
    (2) w/o LW: Removes the learnable global pattern weighting module and relies solely on distance-based local weights.
    (3) w/o PE: Substitutes probabilistic (Beta) embeddings with vector embeddings, and accordingly replaces KL divergence and probabilistic conjunction with cosine similarity and weighted summation.

\begin{figure*}[htbp]
  \centering
  \begin{subfigure}{0.45\textwidth}
    \centering
    \includegraphics[width=\linewidth]{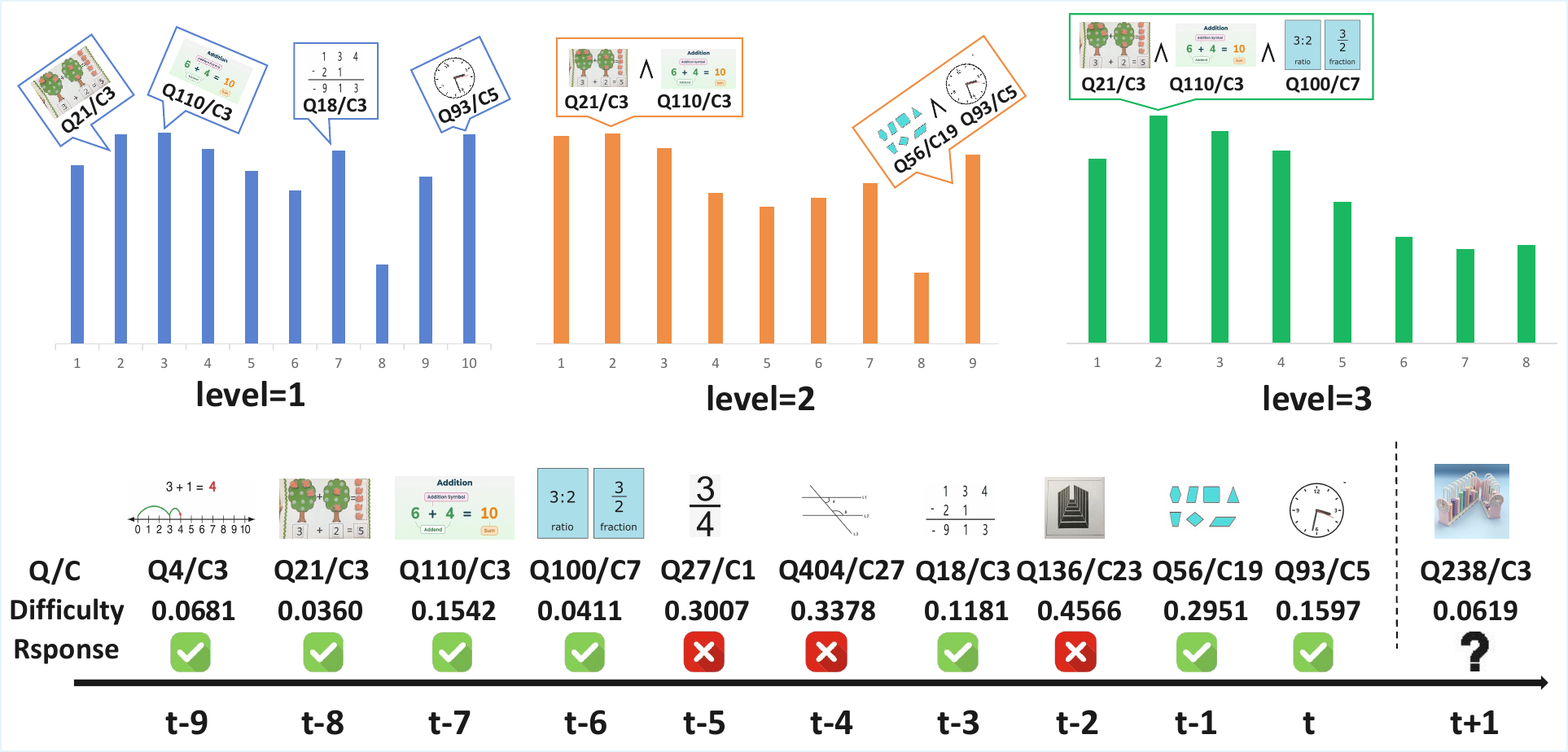}
    \caption{Case study of a real student.
    The upper bar chart shows pattern-level weights, where larger values indicate stronger relevance to the target question.
    ‘$\wedge$’ denotes a conjunction relationship.
    The lower panel presents the student’s historical interaction sequence together with the target question.
    “Q/C” denotes the question and knowledge concept ID.
    }
    \label{fig:sub1}
  \end{subfigure}
  \hfill 
  \begin{subfigure}{0.49\textwidth}
    \centering
    \includegraphics[width=\linewidth]{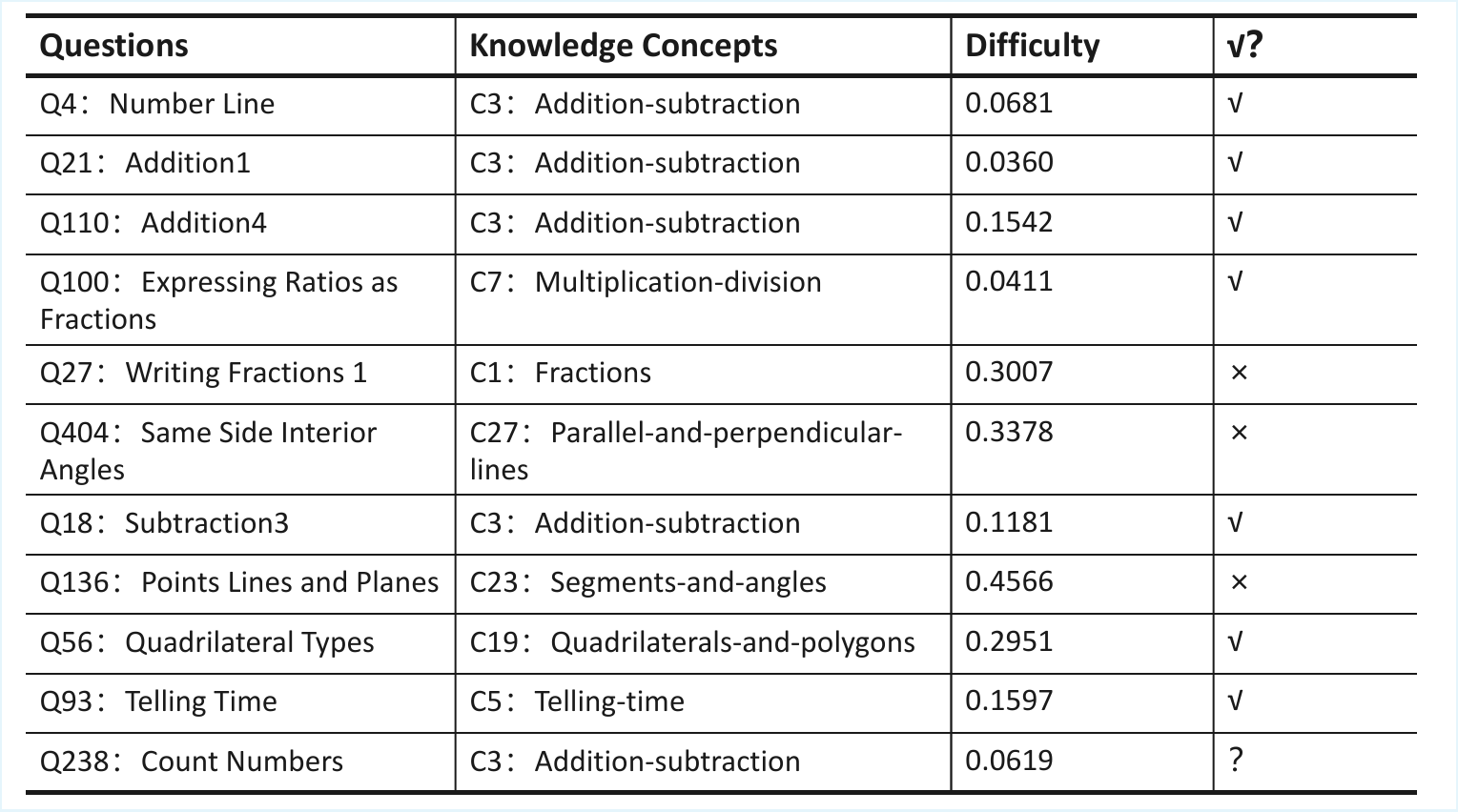}
    \vspace{-.5em}\caption{Expanded view of the interaction history in Figure~\ref{fig:sub1}.
    Each interaction is annotated with the question name, the associated knowledge concept, the question difficulty, and the response outcome.
    Interactions are ordered chronologically from top to bottom.}
    \label{fig:sub2}
  \end{subfigure}
  \vspace{-.5em}\caption{Interpretable case study: distribution of pattern contribution for \emph{correct-responses} on the Junyi Dataset.}
  \label{fig:case_study}
\end{figure*}

Removing the interaction-level conjunction (–w/o Inter) consistently degrades performance across all datasets, indicating that simple concatenation leads to shallow feature fusion and fails to capture the joint constraints between questions and knowledge concepts.
Eliminating the learnable pattern weight module (–w/o LW) causes notable performance drops, in some cases exceeding those caused by removing interaction modeling. 
This highlights the importance of incorporating global sequence information beyond local pattern–target similarity.
Despite these removals, the degraded variants still outperform most baselines, demonstrating the robustness of the overall PLKT framework.
Replacing probabilistic embeddings with vector embeddings (–w/o PE) resulted in the most severe performance degradation (AUC loss up to 17\%), confirming that probabilistic representations are crucial for effective pattern inference and uncertainty modeling.
Probabilistic embeddings inherently support distribution-based logical operations and distance metrics, making the model theoretically more aligned with cognitive modeling needs and enhancing interpretability.
Overall, these results demonstrate that all components are necessary and jointly enable PLKT’s strong performance.

\subsubsection{Analysis of Pattern Level}
PLKT employs a multi‑level pattern mechanism to capture learning behaviors at varying granularities.
Figure~\ref{fig:pattern_level_09_12} analyzes the impact of pattern level on model performance.
Due to space limits, only results on ASSIST09 and Bridge06 are shown in the main text; results on Junyi and ASSIST12 are provided in Supplementary Material B.3.

As the number of pattern levels increases, PLKT consistently achieves improved performance, demonstrating that multi-level modeling substantially enhances representation capacity.
ASSIST09 achieves optimal performance at $level=5$, while Bridge2006 benefits from deeper pattern hierarchies and peaks at $level=6$.
In particular, using multiple levels consistently outperforms the point‑level.
These observations suggest that the multi-level mechanism captures complementary information across different granularities.
Lower-level patterns emphasize local interactions and recent response signals, whereas higher-level patterns capture more stable learning structures.
By jointly modeling multiple levels, PLKT strikes a balance between local sensitivity and global consistency, resulting in more expressive representations of the student.

\subsubsection{Case Study of Explainability}
To illustrate the interpretability of PLKT, we conduct case studies on two representative students from the Junyi dataset.
Since PLKT adopts outcome-conditioned reasoning with distinct pattern scores and weights for \emph{correct} and \emph{incorrect} response states, we analyze its explanations under both scenarios.
Due to space constraints, only the \emph{correct-response} case is shown in the main text, while the \emph{incorrect-response} case is deferred to Supplementary Material B.4.
Figure~\ref{fig:case_study} visually represents the contribution distribution of behavioral patterns at different levels when students provide correct responses.

At $Level=1$ (point-level patterns), PLKT assigns high weights to historically correct responses on conceptually aligned questions (e.g., Q21, Q110, and Q18), reflecting stable mastery of the target knowledge concept.
Q93 (Telling Time) also receives a non-negligible weight due to its implicit reliance on arithmetic operations and its temporal proximity.
In contrast, incorrectly answered questions involving unrelated geometric concepts (e.g., Q404 and Q136) are assigned low weights, indicating effective suppression of irrelevant evidence.
At $Level=2$ (union-level patterns), union patterns formed by consecutive, concept-consistent correct responses (e.g., Q21 and Q110) contribute most strongly.
This demonstrates the short-term knowledge enhancement effect and confirms that PLKT captures the knowledge consolidation effect resulting from consecutive correct responses. 
Conversely, associative patterns with weak conceptual coherence (e.g., Q93 and Q56) are down-weighted, suggesting that PLKT does not rely solely on recency.
At $Level=3$, these arithmetic patterns remain influential when integrated with higher-level but related concepts, revealing longer-term and cross-concept learning structures.
This highlights the complementary roles of different pattern levels.

Overall, PLKT achieves interpretability through explicit multi-level pattern weighting, allowing transparent attribution of predictions to cognitively meaningful learning patterns.
Compared with traditional knowledge tracing models based on opaque hidden states or attention scores, PLKT offers more intuitive and cognitively grounded explanations.

\section{Conclusion}
This paper proposes a novel Probabilistic Logical Knowledge Tracing (PLKT) model that integrates probabilistic embeddings with pattern-based reasoning to achieve accurate and interpretable predictions.
PLKT represents knowledge states using Beta-distributed embeddings and extracts multi-level learning patterns as explicit reasoning units, which are aggregated through probabilistic logical conjunction.
Experiments on five public datasets show competitive or state-of-the-art performance, and case studies demonstrate clear, cognitively grounded interpretability.

\clearpage
\bibliographystyle{named}
\bibliography{ijcai26}

\clearpage
\appendix
\section{NOTATION TABLE}
\label{app:notation_table}
Due to the complicated structure of PLKT, we have listed and explained the relevant symbols in Table \ref{app:tab_expl}.

\begin{table}[htbp]
\centering
\begin{tabularx}{0.5\textwidth}{c|>{\centering\arraybackslash}X}
\toprule
Notation & Description \\
\midrule
$u \in \mathcal{U}$, $q \in \mathcal{Q}$, $c \in \mathcal{C}$ & student, question, knowledge concept \\
$r \in \{0,1\}$ & response of student (1: Correct, 0: Incorrect) \\
$d$ & vector dimension \\
$h_{\mathrm{emb}}^q,h_{\mathrm{emb}}^c$ & question/concept embedding functions \\
$\bm{\alpha},\bm{\beta} \in \mathbb{R}^d$ & parameters of the Beta distribution \\
$(\bm{\alpha}_{t+1}^+,\bm{\beta}_{t+1}^+)\in \mathbb{R}^{2d}$  & distribution of “Correct” status for target questions\\
$(\bm{\alpha}_{t+1}^-,\bm{\beta}_{t+1}^-) \in \mathbb{R}^{2d}$ &  distribution of “Incorrect” status for target questions \\
$l = 1,2,\ldots,L$ & pattern level \\
$L$ & maximum pattern level \\
$\mathbf{p}_k^{(l)}$ & the $k$-th pattern of the $l$-th level \\
$(\bm{\alpha}_k^{(l)},\bm{\beta}_k^{(l)})$ & conjunction distribution parameters of pattern $p_k^{(l)}$ \\
$D^{+(l)}_k,D^{-(l)}_k$ & KL divergence distance between model $p^{(l)}_k$ and the correct/incorrect response target \\
$N_q$ & total number of attempts for question $q$ \\
$M_q$ & number of correct answers for question $q$ \\
$d_q$ & question difficulty  \\
$\eta_k^{+(l)},\eta_k^{-(l)}$ & distance-based weight \\
$\delta_k^{(l)}$ & learnable pattern weights \\
$\gamma$ & margin parameter \\
$\mathbf{u} \in \mathbb{R}^{2d}$ & interaction representation \\
$\mathbf{p} \in \mathbb{R}^{2d}$ & pattern representation \\
\bottomrule
\end{tabularx}
\caption{The notation table of PLKT.}
\label{app:tab_expl}
\end{table}


\begin{figure*}[htbp]
  \centering
  \begin{subfigure}{0.45\textwidth}
    \centering
    \includegraphics[width=\linewidth]{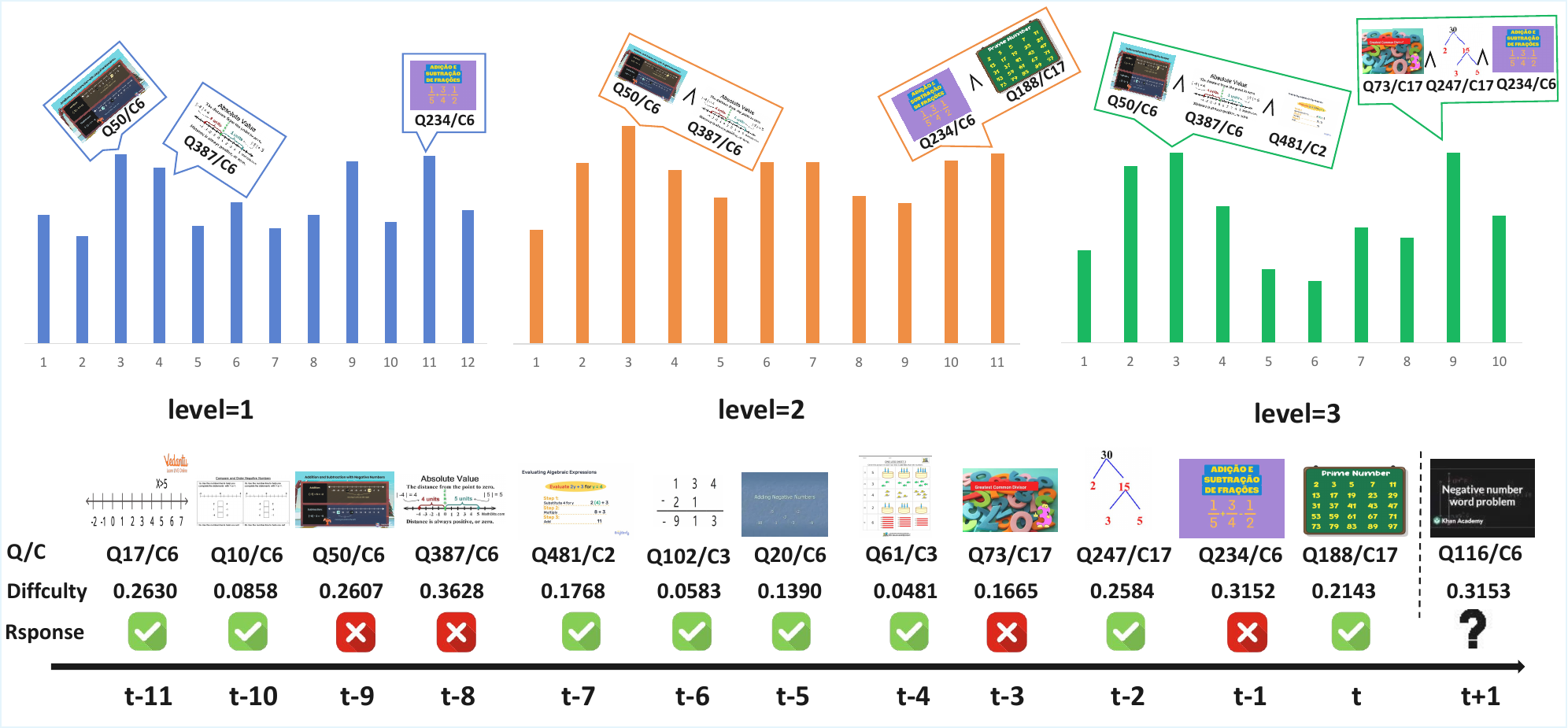}
    \caption{Case study of a real student.
    The upper bar chart shows pattern-level weights, where larger values indicate stronger relevance to the target question.
    ‘$\wedge$’ denotes a conjunction relationship.
    The lower panel presents the student’s historical interaction sequence together with the target question.
    “Q/C” denotes the question and knowledge concept ID.
    }
    \label{app:fig_sub1}
  \end{subfigure}
  \hfill 
  \begin{subfigure}{0.45\textwidth}
    \centering
    \includegraphics[width=\linewidth]{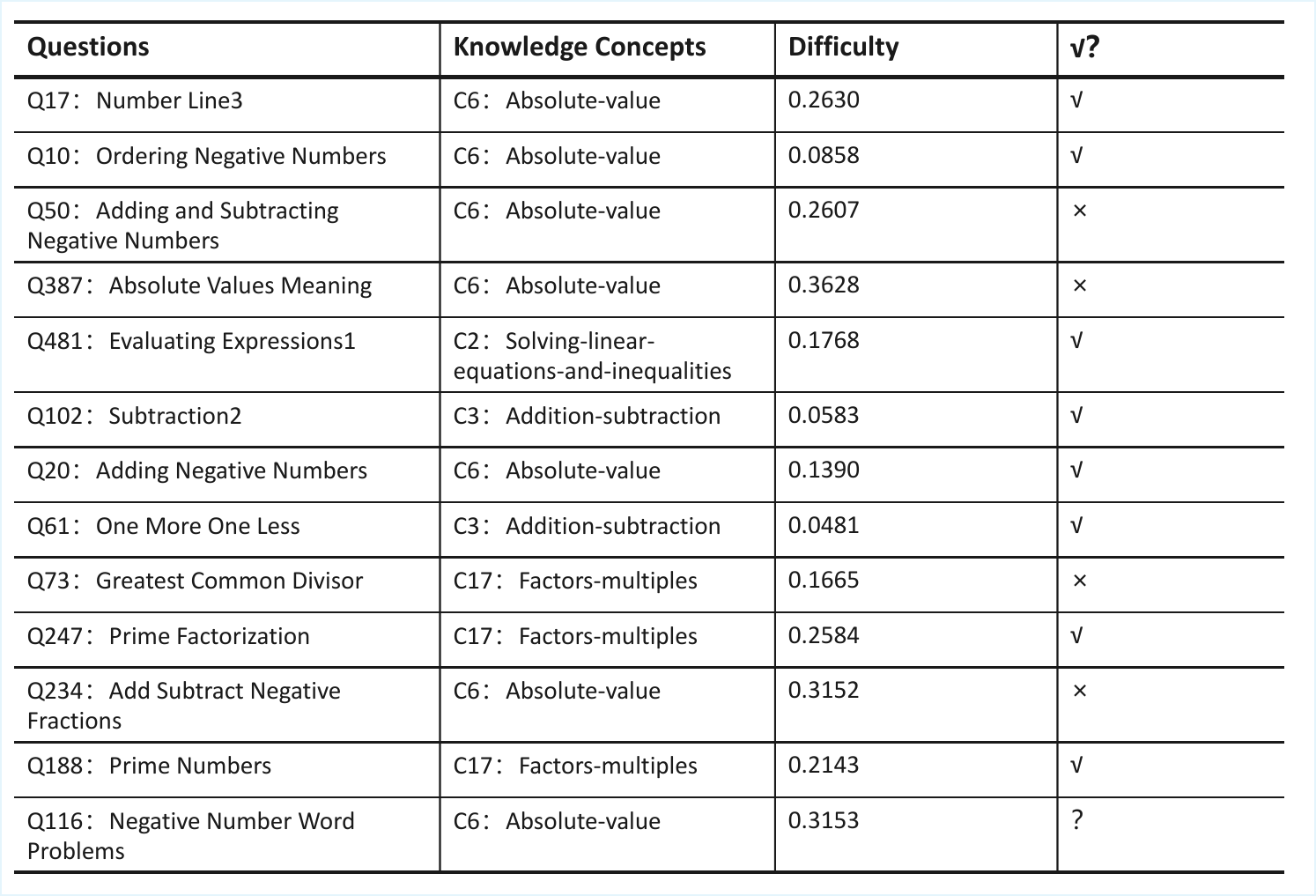}
    \caption{Expanded view of the interaction history in Figure~\ref{app:fig_sub1}.
    Each interaction is annotated with the question name, the associated knowledge concept, the question difficulty, and the response outcome.
    Interactions are ordered chronologically from top to bottom.}
    \label{app:fig_sub2}
  \end{subfigure}
  \caption{Interpretable case study: distribution of pattern contribution for \emph{incorret-responses} on the Junyi Dataset.}
  \label{fig:case_study_case2}
\end{figure*}

\section{SUPPLEMENTS FOR EXPERIMENTS}
\label{app:supp_exercise}
\subsection{Dataset Description}
\label{app:dataset}
To provide a comprehensive description of the experimental data, we conduct evaluations on five representative real-world benchmark datasets widely used in knowledge tracing research. 
These datasets differ substantially in scale, knowledge structure, and interaction density, enabling a thorough assessment of model robustness and generalization.

\begin{itemize}
    \item \textbf{ASSIST2009\footnote{\url{https://sites.google.com/site/ASSISTmentsdata/home/2009-2010ASSISTment-data}}} (ASSIST09):
    ASSIST09 was collected from the ASSISTments online tutoring system during 2009–2010 and primarily focuses on middle-school mathematics learning scenarios.
    We adopt the officially released merged version, which includes student response logs, question identifiers, and annotated knowledge concepts.
    After standard preprocessing, the dataset contains 4,163 students, 17,737 questions, and 458,766 interaction records, covering 123 knowledge concepts.
    Each question is associated with an average of 1.197 concepts.
    ASSIST09 is one of the most commonly used benchmark datasets in knowledge tracing studies.

    \item \textbf{ASSIST2012\footnote{\url{https://sites.google.com/site/ASSISTmentsdata/home/2012-13-school-data-with-affect}}} (ASSIST12):
    ASSIST12 is another large-scale dataset collected from the ASSISTments platform between 2012 and 2013.
    It consists of 27,145 students, 50,966 questions, 245 knowledge concepts, and 2,627,118 interaction records.
    Owing to its scale and dense interaction structure, ASSIST12 is frequently used to evaluate the scalability and performance of knowledge tracing models in large learning environments.

    \item \textbf{Junyi\footnote{\url{https://pslcdatashop.web.cmu.edu/Files?datasetId=1198}}}:
   The Junyi dataset is collected from the Junyi Academy online learning platform (2015) and has been widely used in knowledge tracing research.
    Due to its large scale, we randomly sample a balanced subset of 500,000 interactions for computational efficiency, involving 87,527 students, 705 questions, and 39 knowledge concepts.

    \item \textbf{Algebra2005\footnote{\url{https://pslcdatashop.web.cmu.edu/KDDCup/}}} (Algebra05):
    Algebra05 is a classical dataset released as part of the KDD Cup 2010 Educational Data Mining Challenge, capturing student learning processes in algebra courses.
    It is extensively used in student modeling and knowledge state prediction research.
    The dataset contains 607,025 interaction records, 173,113 questions, and 112 knowledge concepts, with each question linked to an average of 1.363 concepts.

    \item \textbf{Bridge2006} (Bridge06):
    Bridge06 is also sourced from the KDD Cup 2010 Educational Data Mining Challenge.
    It comprises 1,817,476 interaction records generated by 1,146 students, covering 129,263 questions and involving 493 knowledge concepts.
    Compared with other datasets, Bridge06 exhibits higher conceptual granularity and structural complexity.
\end{itemize}

\begin{figure}
    \centering
    \includegraphics[width=\linewidth]{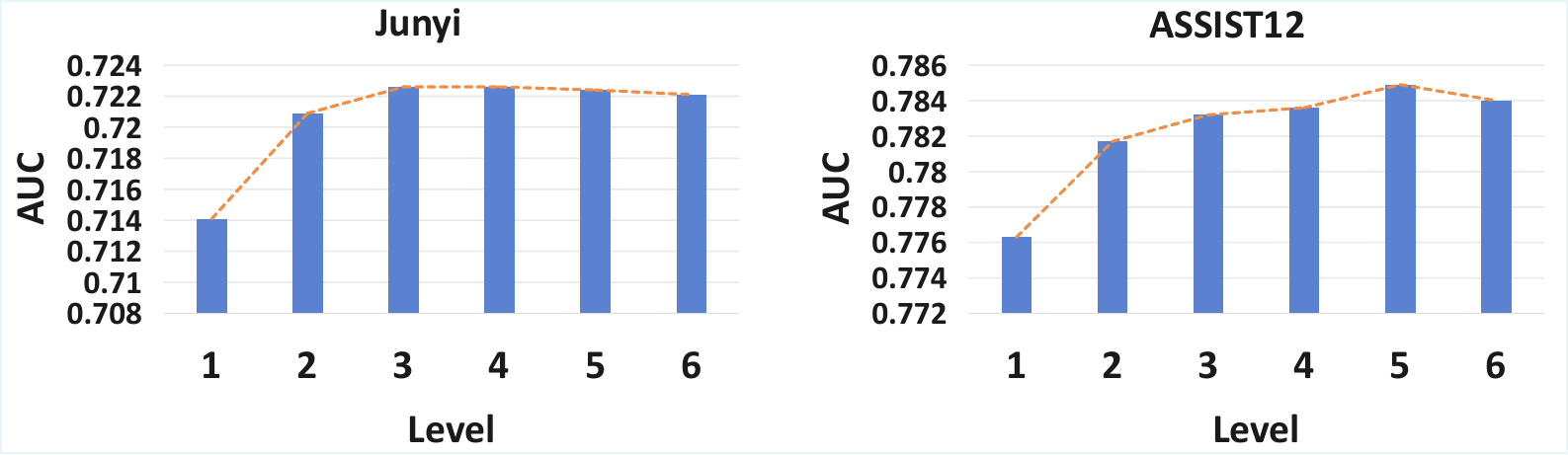}
    \vspace{-1em}\caption{AUC performance of PLKT with varying pattern-levels on Junyi and ASSIST12. $level=1$ denotes point level, while $level>1$ indicates simultaneous use of multiple levels.}
    \label{fig:pattern_level_junyi_bri06}
\end{figure}

\subsection{Baselines}
\label{app:baseline}
To evaluate the effectiveness of the proposed PLKT model, we compare it with twelve representative knowledge tracing baselines published between 2015 and 2025, covering recurrent, memory-augmented, attention-based, psychometric-inspired, and recent uncertainty-aware or interpretable KT models. 
\begin{itemize}
\item DKT: A seminal deep knowledge tracing model that employs recurrent neural networks (RNNs) to dynamically evolve students’ knowledge states. It established a strong performance baseline for subsequent deep learning‑based KT methods.
\item DKT+: An enhanced variant of DKT that introduces consistency‑constraint losses to alleviate prediction inconsistencies on unseen knowledge concepts, thereby improving training stability.
\item DKVMN: A dynamic key‑value memory network that maintains a static key matrix (concepts) and a dynamic value matrix (mastery levels), updating knowledge states through dedicated read‑write operations.
\item SAKT: The first KT model that adopts the self‑attention mechanism. It captures long‑range dependencies across interactions by relating past exercises to the target, overcoming the limited long‑term modeling ability of RNNs.
\item AKT: A Transformer‑based model that integrates item response theory (IRT) with a monotonic attention mechanism. It uses Rasch‑model embeddings and regularization to achieve both interpretability and high predictive accuracy.
\item SAINT: A full Transformer architecture that separately encodes exercise information (encoder) and response information (decoder), leveraging deep self‑attention to model complex exercise‑answer relationships.
\item SimpleKT: A lightweight attention‑based model that explicitly incorporates exercise‑specific features and dot‑product attention to distinguish among similar exercises, balancing performance and efficiency.
\item DeepIRT: A hybrid model that combines the memory‑based tracking of DKVMN with item response theory, providing a psychometric interpretation of student ability and item difficulty.
\item CSKT: A model designed to address the cold‑start problem. It employs kernel bias for short‑sequence modeling and a conical attention mechanism to capture hierarchical knowledge structure.
\item LPKT: A learning‑process‑aware model that takes (exercise, response time, answer) triples as input, uses a learning gate to model knowledge absorption and a forgetting gate to simulate knowledge decay.
\item SparseKT: A sparse‑attention framework that selects critical interactions via a k‑selection module and a soft‑threshold sparse attention mechanism, reducing overfitting and enhancing robustness.
\item UKT: An uncertainty‑aware model that represents each interaction with a random distribution and employs a Wasserstein self‑attention mechanism to track the evolution of knowledge state distributions.
\end{itemize}

\begin{figure*}[t]
    \centering
    \begin{subfigure}[b]{0.24\textwidth}
        \centering
        \includegraphics[width=\textwidth]{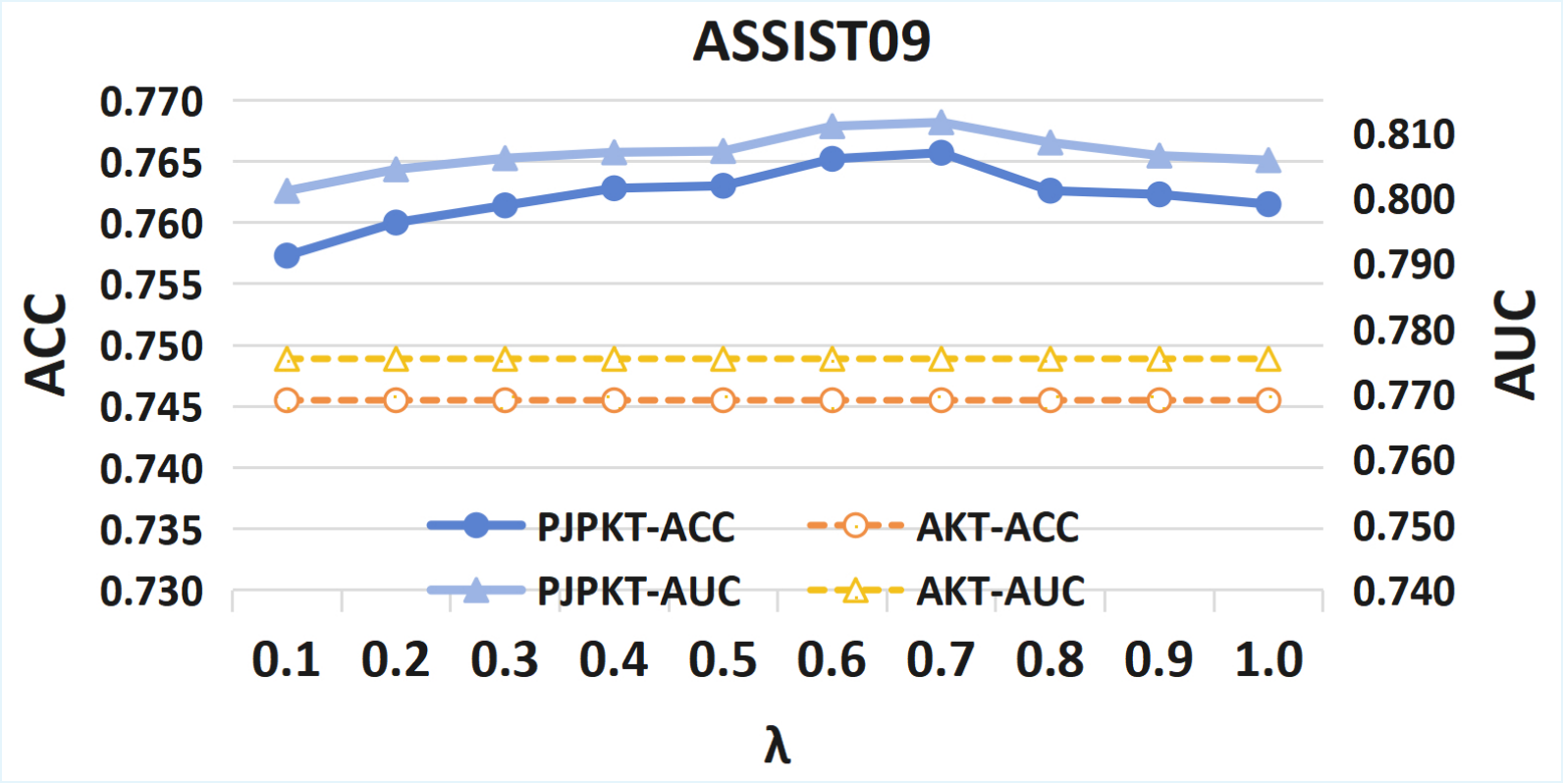}
        \caption{ASSIST09}
        \label{fig:lambda_ASSIST09}
    \end{subfigure}\hfill  
    \begin{subfigure}[b]{0.24\textwidth}
        \centering
        \includegraphics[width=\textwidth]{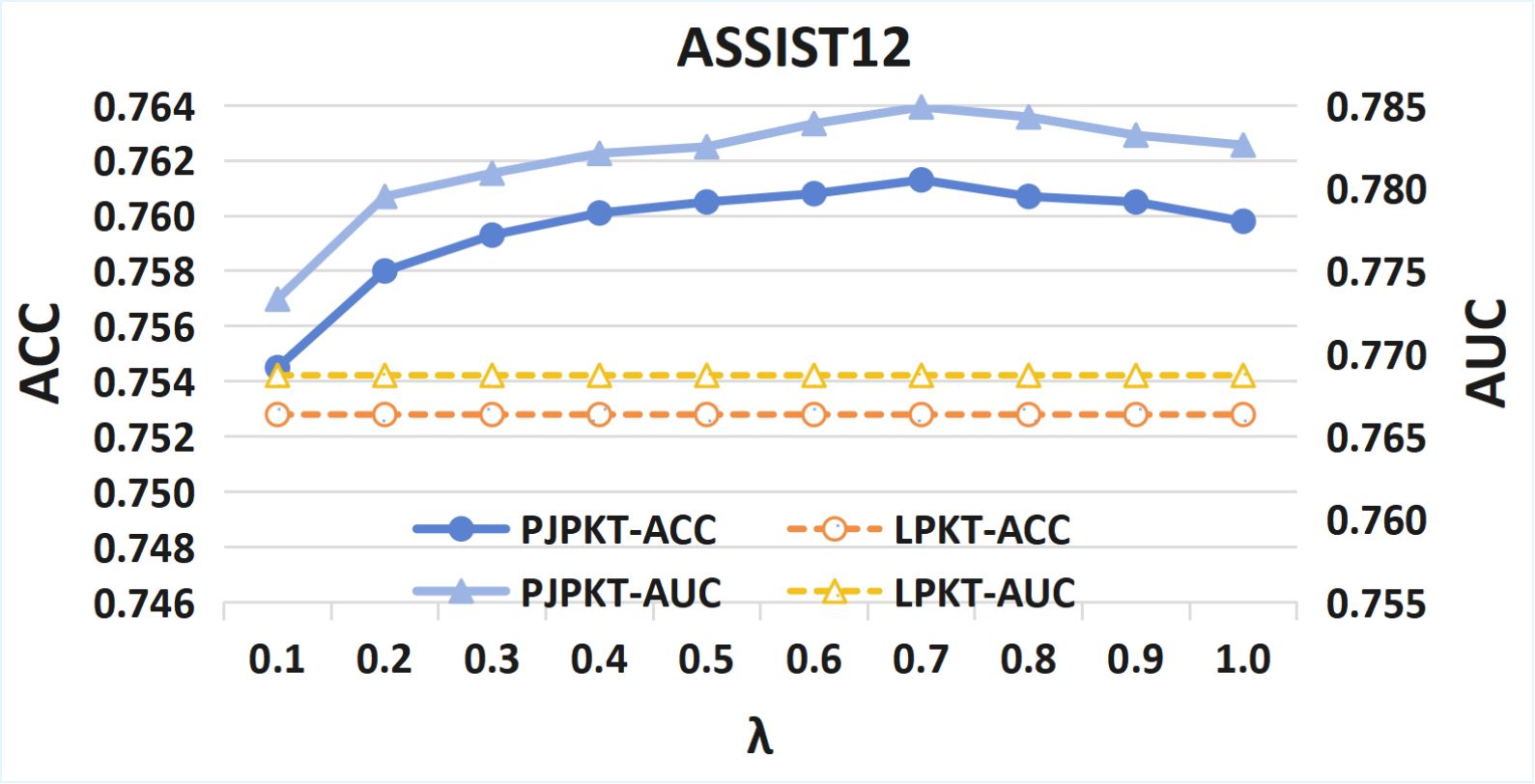}
        \caption{ASSIST12}
        \label{fig:lambda_ASSIST12}
    \end{subfigure}\hfill  
    \begin{subfigure}[b]{0.24\textwidth}
        \centering
        \includegraphics[width=\textwidth]{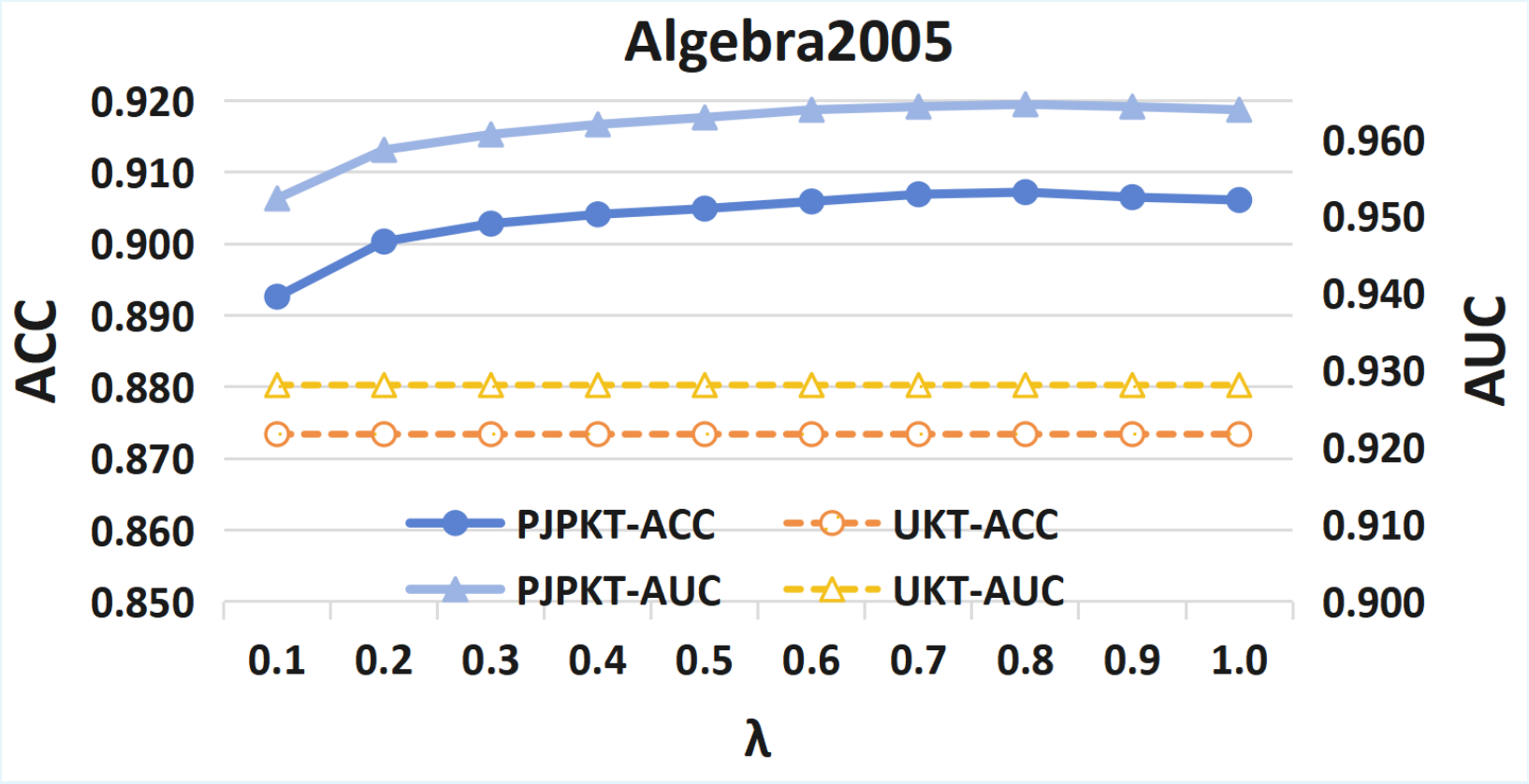}
        \caption{Algebra 2005}
        \label{fig:lambda_algebra2005}
    \end{subfigure}\hfill  
    \begin{subfigure}[b]{0.24\textwidth}
        \centering
        \includegraphics[width=\textwidth]{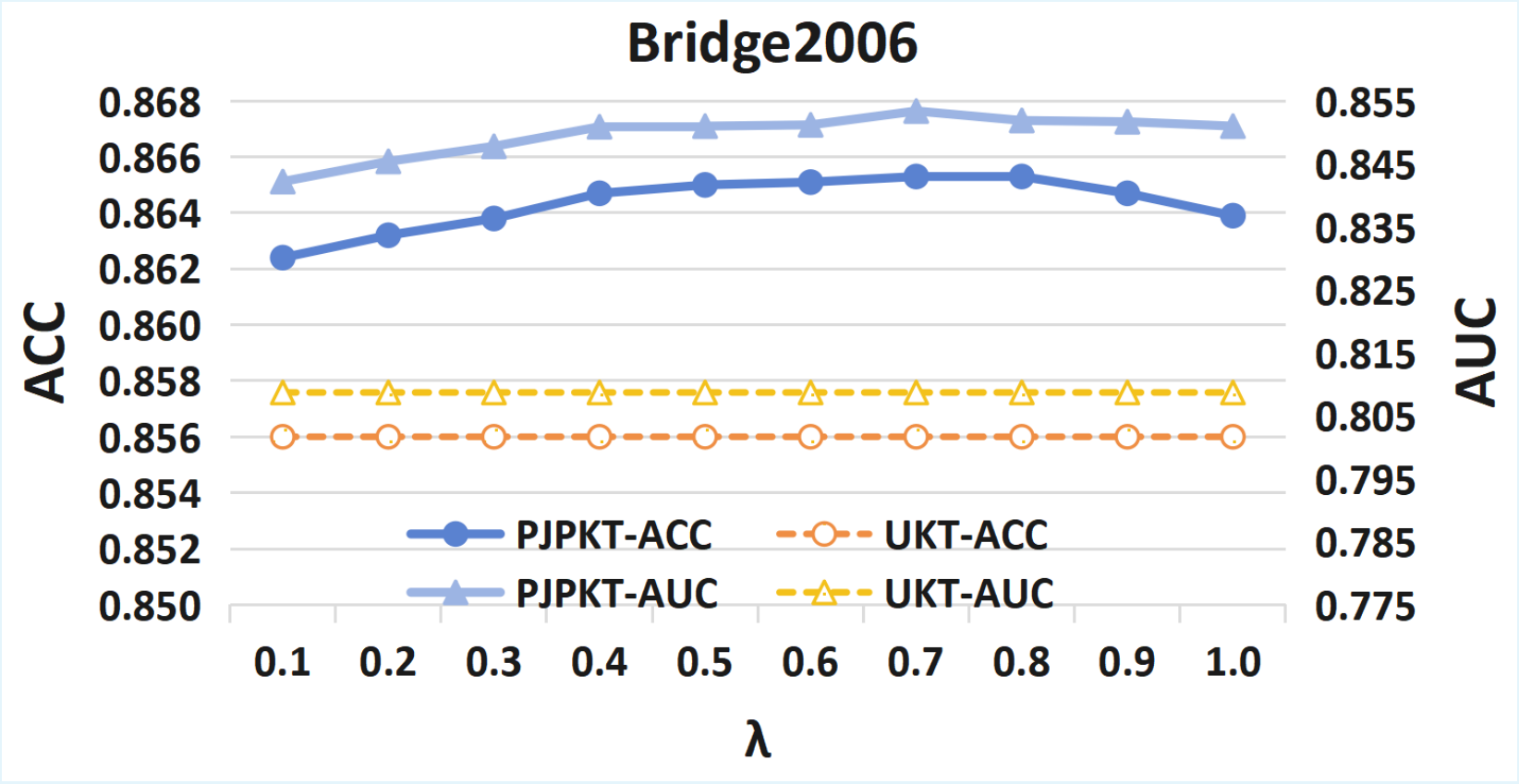}
        \caption{Bridge 2006}
        \label{fig:lambda_bridge2006}
    \end{subfigure}
    \caption{Sensitivity analysis of hyperparameter $\lambda$ across four benchmark datasets.}
    \label{fig:lambda_sensitivity}
\end{figure*}

\subsection{Extended Analysis of Pattern‑Level}
This section reports the complete pattern‑hierarchy results on the Junyi and ASSIST12 datasets (Figure~\ref{fig:pattern_level_junyi_bri06}). Performance improves steadily as the number of pattern levels increases on both datasets. ASSIST12 achieves optimal performance at $level=5$, while Junyi attains its optimum earlier at $level=3$ with diminishing returns thereafter, likely due to its shorter sequences and predominantly single‑concept questions, which reduce the utility of higher‑order joint patterns. 
Consistent with the main text, multi-level modeling consistently surpasses point-level modeling across all datasets, confirming the advantage of integrating multiple levels.

\subsection{Case Study of Explainability}
\label{app:case2}

Figure \ref{fig:case_study_case2} shows the pattern-level importance distribution when PLKT predicts an incorrect response to target Q116 (negative number word problems, concept: absolute value).
At $level=1$, the model assigns high negative importance to conceptually aligned questions (e.g., Q50, Q387, and Q234) that were repeatedly answered incorrectly and exhibit comparable difficulty, indicating persistent deficiencies in the target concept and providing strong negative evidence for the prediction.
In contrast, concept-related but correctly answered or temporally distant exercises (e.g., Q17 and Q10) receive lower weights, reflecting effective suppression of weak or contradictory evidence.

At higher levels, union patterns formed by multiple incorrect responses, particularly those involving Q50, retain high importance across $level=2$ and $level=3$, demonstrating stable negative evidence across temporal granularities.
Incorporating algebraically related Q481 further strengthens these patterns, while combinations including weakly related correct responses are down-weighted.
This behavior confirms that PLKT selectively aggregates coherent failure patterns and suppresses inconsistent evidence, yielding interpretable explanations for incorrect predictions.

\subsection{Hyper-parameter Analysis}
\label{app:hyper_para_anal}

We analyze the effect of the weighting parameter $\lambda$, which controls the trade-off between distance-based pattern-target similarity and learnable global pattern weights.
Experiments are conducted on ASSIST09, ASSIST12, Algebra2005, and Bridge2006 with $\lambda \in [0,1]$.
Figure~\ref{fig:lambda_sensitivity} reports the performance trends, where the best baseline result on each dataset is shown for reference.

Across all datasets, performance exhibits a consistent trend of initial improvement followed by a mild decline as $\lambda$ increases.
ASSIST09, ASSIST12, and Bridge2006 achieve optimal performance at $\lambda=0.7$, while Algebra2005 peaks at $\lambda=0.8$.
Notably, PLKT consistently outperforms the strongest baseline under all $\lambda$ settings.

Moderate values of $\lambda$ enhance performance by enabling global pattern weights to complement local similarity signals, allowing the model to better capture long-range dependencies and sequence-level learning trends.
This effect is particularly pronounced on datasets with longer interaction sequences and richer behavioral structures, such as ASSIST12 and Bridge2006.
In contrast, excessively large $\lambda$ values overemphasize global weights and weaken fine-grained pattern-target alignment, leading to performance degradation.

Overall, PLKT shows low sensitivity to $\lambda$ and maintains stable performance over a wide range ($0.5 \leq \lambda \leq 0.9$), demonstrating robust parameter behavior.
These results validate the effectiveness of balancing local similarity modeling with global pattern importance weighting.


\end{document}